\documentclass[letterpaper]{article} 
\usepackage{aaai2026}  
\usepackage{times}  
\usepackage{helvet}  
\usepackage{courier}  
\usepackage[hyphens]{url}  
\usepackage{graphicx} 
\usepackage{natbib}  
\usepackage{caption} 
\usepackage{amsmath}
\usepackage{amssymb}
\usepackage{booktabs}
\usepackage{multirow}
\usepackage{xcolor}
\usepackage{pifont}
\usepackage{makecell}
\definecolor{cYes}{HTML}{16A34A}
\definecolor{cNo}{HTML}{DC2626}
\definecolor{cPartial}{HTML}{D97706}

\usepackage{algorithm}
\usepackage{algorithmic}

\usepackage{newfloat}
\usepackage{listings}
\DeclareCaptionStyle{ruled}{labelfont=normalfont,labelsep=colon,strut=off} 
\floatstyle{ruled}
\newfloat{listing}{tb}{lst}{}
\floatname{listing}{Listing}
\title{OIL: An Object Interaction Layer for Humanoid Whole-Body Control}
\author{
    Pranay Dugar, 
    Mohitvishnu S. Gadde, 
    Rajesh Mangannavar, \\
    Aayam Kumar Shrestha, 
    Alan Fern
}

\affiliations{
    \text{{dugarp, gaddem, mangannr, shrestaa, afern}@oregonstate.edu}\\
    Collaborative Robotics and Intelligent Systems Institute (CoRIS)\\
    Oregon State University\\
    Corvallis, Oregon 97331, USA
}

\begin{document}

\maketitle

\begin{abstract}

Recent progress in learned whole-body controllers for humanoid robots has enabled robust free-space motion by tracking kinematic targets such as joint or end-effector trajectories. This interface provides a useful abstraction for upstream systems, including motion capture, teleoperation, or generative models, which can specify desired motion without reasoning about forces or balance. However, this abstraction has largely been limited to robot motion, requiring task-specific controllers or retraining for object interaction. We introduce OIL, an Object Interaction Layer that extends the kinematic command interface of whole-body control to include desired object motion and contact. OIL augments a standard controller with object-centric commands and translates them into coordinated whole-body behavior while preserving locomotion and balance. This design enables a single learned policy to execute non-prehensile interactions, including pushing, pulling, two-arm pick-and-place, and locomotion-coupled transport, directly from object-level kinematic targets without task-specific retraining. In simulation, we demonstrate that object-level commands improve performance over robot-centric controllers while matching task-specific policies across variations in object properties such as mass and size. We further demonstrate zero-shot generalization to novel behavior combinations and sim-to-real transfer on the Digit humanoid robot.
\end{abstract}


\section{Introduction}

\begin{figure}[t]
  \centering
  \includegraphics[width=\linewidth]{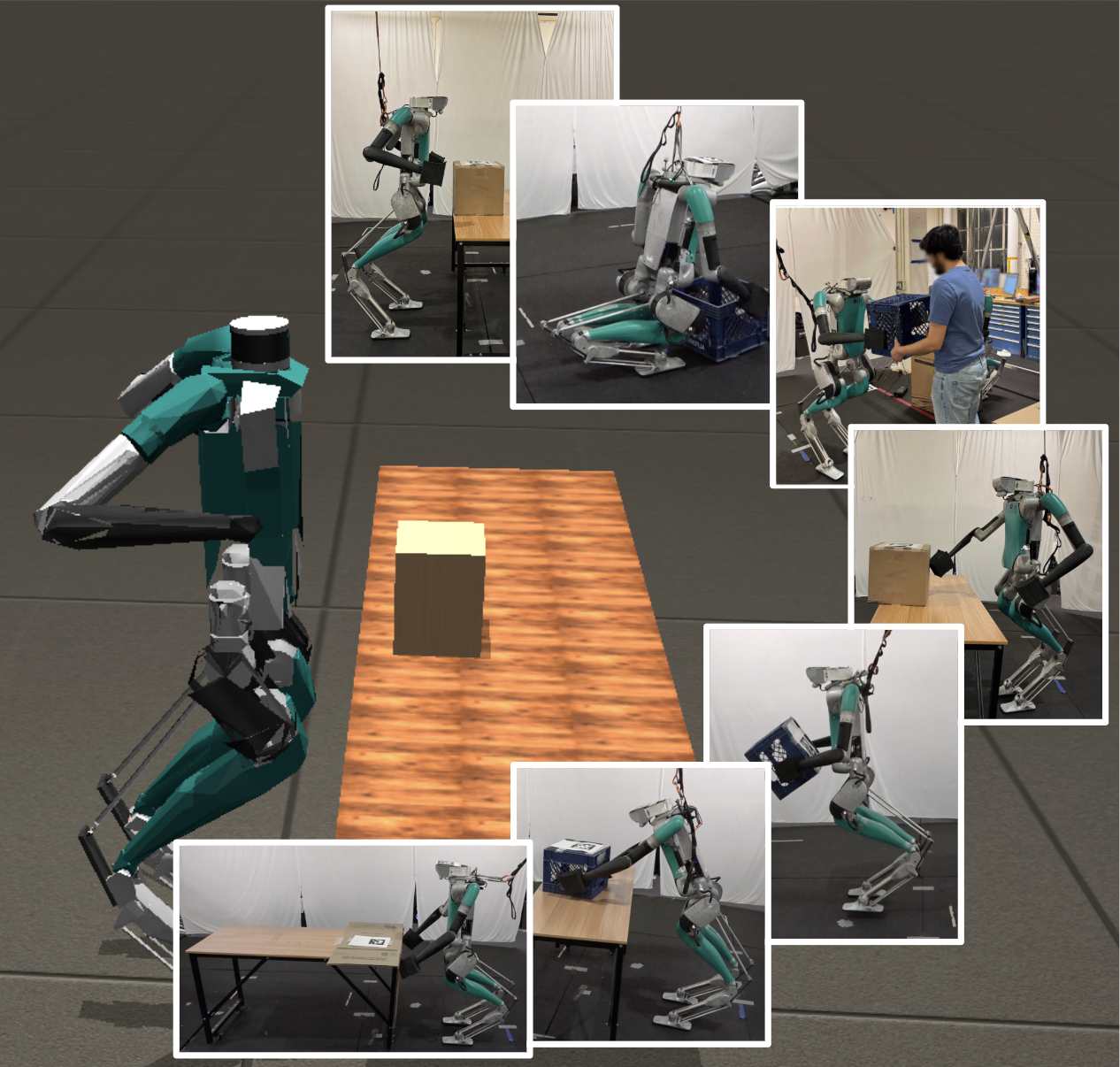}
  \caption{Overview of OIL. An object-interaction layer that augments a whole-body controller with object-level motion and contact directives, enabling a single controller to execute diverse loco-manipulation behaviors such as walking up to a box, picking it up from different heights and positions in front of the robot, walking while holding it up, placing it down at different locations, and pushing it on the surface with one or two arms. It further allows composition of these skills without any per-task finetuning.}
  \label{figure:title}
\end{figure}

Humanoid robots deployed in homes, warehouses, and industrial environments are expected to combine locomotion with object interaction within a single continuous task. A robot may need to walk to a shelf, push or slide an object, lift it with two arms, carry it while navigating obstacles, and reposition items along the way. Such behaviors require coordinated whole-body motion under changing contact conditions. Importantly, many of these tasks can be specified at a high level through kinematic targets that describe the desired motion of both the robot and the manipulated object. This abstraction allows upstream systems, including planners, authoring tools, motion capture, or perception modules, to specify intended motion without explicitly reasoning about contact forces or balance strategies. Humans naturally operate at this level, translating desired object and body motion into coordinated whole-body behavior without consciously computing forces.

Recent advances in learning-based humanoid control have shown that kinematic command interfaces are a powerful and general abstraction for robot motion. Modern whole-body controllers (WBCs) such as HOVER \cite{he2025hover}, HumanPlus \cite{fu2024humanplus}, and the Masked Humanoid Controller (MHC) \cite{dugar2025learning} accept joint or end-effector trajectory targets and reliably produce coordinated locomotion, balance, and upper-body motion within a single policy. These systems demonstrate that a unified controller can robustly track diverse robot motion targets without retraining for each behavior, reducing the need for explicit mode switching and enabling a broad range of coordinated whole-body behaviors.

However, this kinematic abstraction has largely been limited to robot body kinematics, without explicit intent for object interaction. Most general-purpose whole-body controllers remain robot-centric, driven by joint targets, end-effector poses, or reference trajectories with no explicit representation of intended object motion or contact. Yet appropriate whole-body behavior depends critically on these object-level intents: pushing, lifting, and pulling demand different force generation, contact timing, and balance strategies, even when the robot's kinematics appear similar. Without object-aware inputs, interaction intent must be inferred implicitly from robot motion, limiting robustness and generalization.

Recent work has begun incorporating object state and contact information directly into whole-body control. Hierarchical and residual approaches condition policies on object trajectories, contact signals, or target forces to enable coordinated locomotion and manipulation \cite{liu2025opt2skill, zhao2025resmimic, tessler2025maskedmanipulator, zhang2024wococo}. While these methods improve performance, even with limited demonstrations, they are typically trained for specific tasks, with separate controllers learned for each behavior. Even closely related variants, such as pushing the same object along a different path, often require retraining. As a result, humanoid systems still lack a reusable object-interaction abstraction within whole-body control.

The goal of this work is to enable humanoid robots to reliably execute object-interaction behaviors specified through kinematic targets on both the robot body and the object pose. In this formulation, high-level systems can augment robot body motion with object motion and contact, without reasoning about task-specific force generation or balance. To support this abstraction, we introduce OIL, an Object Interaction Layer that augments a standard whole-body controller with object motion and contact commands. OIL translates these object-level goals into coordinated whole-body behavior while preserving the locomotion and balance capabilities of the base controller. This design enables a single learned policy to perform pushing, pulling, two-arm pick-and-place, and locomotion-coupled transport without task-specific retraining.

Our contributions are as follows:\\
{\bf Object-Intent API for Whole-Body Control.} We propose a simple extension to kinematic whole-body directives that exposes desired object motion and contact intent as first-class command inputs for humanoid WBCs.\\
{\bf OIL Architecture and Training.} We propose the OIL architecture, which minimally extends a frozen WBC by generating upper-body directives from object intent, and a corresponding data generation and curriculum-based training procedure that supports a single policy over diverse interactions.\\
{\bf OIL Evaluation.} We evaluate OIL across a range of non-prehensile object-interaction behaviors and object properties, and show generalization to interactions beyond those used during training, including a sim-to-real demonstration on the Digit humanoid robot.

\begin{figure*}[t]
  \centering
  \includegraphics[width=0.9\textwidth]{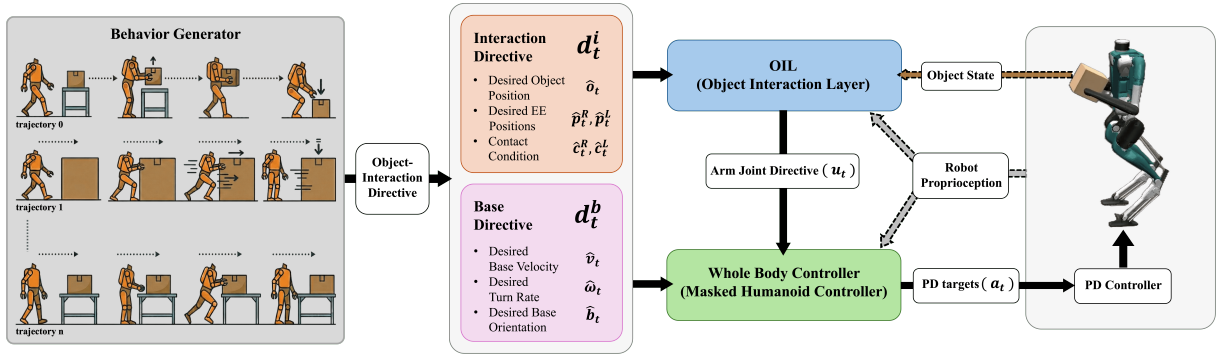}
  \caption{\textbf{Architecture of the Object-Interactive Controller}. A behavior generator produces an object-interaction directive, split into an interaction directive $d^i_t$ (desired object position $\hat{o}_t$, end-effector targets $\hat{p}^R_t$, $\hat{p}^L_t$, and contact conditions $\hat{c}^R_t$, $\hat{c}^L_t$) and a base directive $d^b_t$ (desired velocity $\hat{v}_t$, turn rate $\hat{\omega}_t$, and base orientation $\hat{b}_t$). OIL (blue), running at 50\,Hz, maps $d^i_t$, object state (updated at 10\,Hz), and robot proprioception to an arm joint directive $u_t$. The Masked Humanoid Controller (MHC, green), also running at 50\,Hz, combines $u_t$ and $d^b_t$ with robot proprioception to produce PD targets $a_t$ for all actuated joints, which are executed by a PD controller at 2\,kHz. During training, the MHC weights remain frozen and only the OIL weights are optimized.}
  \label{figure:architecture}
\end{figure*}

\section{Related Work}
\label{sec:related_work}

\textbf{Whole-body Humanoid Control.} Reinforcement learning has enabled stable bipedal locomotion over rough terrain and under disturbances~\cite{radosavovic2024learning, gadde2025no, li2025reinforcement, van2024revisiting, haarnoja2024learning, rudin2022learning, dugar2025no}, while motion imitation from human data has extended control to expressive whole-body behaviors~\cite{peng2018deepmimic, peng2021amp, luo2023perpetual}. Recent whole-body controllers unify these capabilities within a single policy, supporting joystick commands, end-effector targets, and full-body motion references interchangeably~\cite{dugar2025learning, he2025hover, cheng2024expressive, ji2024exbody2}. These interfaces can also be driven by teleoperation, where a human generates object-interaction trajectories using visual feedback. However, they remain strictly robot-centric, with object intent encoded only implicitly through robot motion. Deploying them for object interaction, therefore, requires translating object-level goals into robot trajectories, which limits the controller’s ability to adapt to deviations in object motion or changes in object dynamics.

\textbf{Trajectory-Guided Loco-Manipulation.} A growing body of work addresses object interaction by generating reference trajectories that explicitly encode robot–object motion, contact timing, and in some cases target forces, and then training policies to track these trajectories at run time. Opt2Skill~\cite{liu2025opt2skill} uses full-order trajectory optimization to produce dynamically feasible whole-body references for Digit, including reference forces. Alternatively, ResMimic~\cite{zhao2025resmimic} learns a residual correction policy on top of a frozen motion tracker using paired human–object motion capture data. Other methods obtain references through teleoperation~\cite{he2024learning, ze2025twist} or video imitation~\cite{fu2024humanplus, weng2025hdmi}, and train autonomous controllers to track the collected demonstrations robustly. In all cases, object motion and contact objectives are embedded within task-specific reference trajectories. As a result, each new interaction requires a dedicated round of data collection, trajectory generation, environment modeling, and reinforcement learning.

\textbf{Task-Specific Object Interaction.} Tasks requiring tighter coupling between locomotion and contact have been addressed by end-to-end RL, though each behavior typically demands dedicated reward engineering and contact scheduling. WoCoCo~\cite{zhang2024wococo} decomposes tasks into contact-stage sequences with dense contact rewards and curiosity bonuses, demonstrating parkour, wall climbing, and box carrying, but each task needs its own reward weights, contact plan, and policy. Dao et al.~\cite{dao2024sim} and Xie et al.~\cite{xie2023hierarchical} similarly train dedicated policies for box loco-manipulation with task-specific state machines. These methods demonstrate that coordinated locomotion and manipulation are achievable, but at the cost of per-task reward design, contact scheduling, and a full training run for every new behavior.

\textbf{Whole-Body Manipulation for Simulated Characters.} Several recent methods target multi-behavior object interaction within a single policy, but do so on high-DoF simulation characters with no real-hardware transfer. MaskedManipulator~\cite{tessler2025maskedmanipulator} distills a motion tracker trained on the GRAB~\cite{taheri2020grab} dataset into a goal-conditioned policy accepting sparse spatiotemporal constraints on both body parts and object pose. It handles multiple behaviors within a single policy. Other methods~\cite{tessler2024maskedmimic, xu2025intermimic, yu2025skillmimic} similarly target diverse interaction behaviors from human demonstrations or goal-conditioned generation. 
While these systems demonstrate impressive multi-behavior control in simulation, they have not yet been demonstrated on physically realistic humanoid platforms or under hardware constraints.

\textbf{Summary.} Existing whole-body controllers are either robot-centric, lacking explicit object representation, or object-aware but trained for specific tasks or trajectory classes. In trajectory-conditioned and residual approaches, object motion and contact objectives are embedded in task-specific references, and policies are trained to track them. As a result, each new interaction or trajectory family requires dedicated training. In contrast, OIL exposes desired object motion and contact as reusable runtime commands within the controller itself. A single learned policy can therefore generalize across diverse object-motion trajectories at run time, without per-task retraining, while preserving the balance and robustness of the underlying whole-body controller.

\section{Problem Formulation}

We consider the problem of learning a humanoid controller capable of executing a broad distribution of object-interaction behaviors specified through kinematic directives. We focus on non-prehensile bimanual end-effector interaction, where the robot manipulates objects using coordinated whole-body motion and contact without dexterous grasping. These behaviors include pushing, pulling, lifting, carrying, and locomotion-coupled transport, with variations in object motion, contact timing, and base movement. The distribution of interaction behaviors may be generated from authored generators, motion capture, teleoperation, video-based reconstruction, or other upstream systems that produce desired robot and object motion targets.

To support this setting, we adopt a directive-based formulation. A directive specifies the intended kinematic outcome of an interaction between a humanoid and an object over a finite horizon, including desired object motion, hand trajectories, and base motion, while leaving force generation and full-body coordination to the controller.

\textbf{Object-Interaction Directives.} We consider a humanoid with $J$ actuated joints and denote the \emph{robot proprioceptive state} at time $t$ by
$r_t = (\theta_t, \dot{\theta}_t, \alpha_t, q_t)$,
where $\theta_t \in \mathbb{R}^J$ are joint positions, $\dot{\theta}_t \in \mathbb{R}^J$ are joint velocities, 
$\alpha_t \in \mathbb{R}^3$ is the base angular velocity, and $q_t \in \mathbb{R}^4$ is the base orientation. We do not include the base linear velocity in $r_t$ due to challenges in reliably estimating it on real hardware. We denote the \emph{object state} at time $t$ by $o_t$, which is an encoding of the object pose in the robot frame.

An \emph{object-interaction directive} over a time horizon $H$ is specified as $d = (d^i_{1:H}, d^b_{1:H})$, where $d^i_t$ and $d^b_t$ are the interaction and (robot) base directives, respectively, at time $t$. The \emph{interaction directive} is specified by $d^i_t = (\hat{o}_t, \hat{p}^R_t, \hat{p}^L_t, \hat{c}^R_t, \hat{c}^L_t)$ where:
\begin{itemize}
    \item $\hat{o}_t \in \mathbb{R}^3$ is the desired object position,
    \item $\hat{p}^R_t, \hat{p}^L_t \in \mathbb{R}^3$ are desired right and left end-effector positions expressed in the base frame,
    \item $\hat{c}^R_t, \hat{c}^L_t \in \{0,1\}$ specify the desired contact condition between the object and right and left end-effectors.
\end{itemize}
The \emph{base directive} is specified by $d^b_t = (\hat{v}_t, \hat{\omega}_t, \hat{b}_t)$, where $\hat{v}_t$ denotes desired base planar velocity, $\hat{\omega}_t$ desired turn rate, and $\hat{b}_t$ desired base orientation parameters.

\textbf{Controller Objective.}
Importantly, an object-interaction directive $d$ does not specify joint torques, contact forces, or whole-body trajectories. Thus, the object-interactive controller, denoted by $\pi^{\textrm{OI}}$, must realize the kinematic constraints encoded in $d$ while filling in unspecified degrees of freedom, such as lower-body coordination and force generation. 

At each time step $t$, the controller receives the current proprioceptive robot state $r_t$, object state $o_t$, and directive $d_t$, which specifies the robot-object target for the next step. The output action is
\[
a_t = \pi^{\mathrm{OI}}(r_t, o_t, d_t),
\]
corresponding to PD setpoints for the $J$ actuated joints. Note that the controller observation does not contain explicit information about end-effector contact due to the lack of contact sensing on the hardware used in this work. The training objective is to produce joint robot–object motions whose realized object motion, end-effector trajectories, and base motion align with the specified directive over the horizon $H$. 

\section{Object-Interactive Whole-Body Control}

We now describe how the directive formulation introduced in the previous section is realized in practice. Our approach builds an object-interactive controller by layering an Object-Interaction Layer (OIL) on top of an existing, well-trained whole-body controller (WBC) trained for free-space (non-interactive) motion. In this work, we use the Masked Humanoid Controller (MHC) as the underlying WBC due to its robustness, support for partially specified kinematic directives, and demonstrated sim-to-real transfer capabilities. The resulting system preserves the stability and locomotion performance of the base controller while extending its command interface to support object-interaction directives.

\subsection{Architecture Overview}

Figure~\ref{figure:architecture} illustrates the overall architecture for the object-interactive controller $\pi^{\textrm{OI}}$. At each time step, the system receives the robot proprioceptive state $r_t$, object state $o_t$, and a directive target $d_t = (d^i_t, d^b_t)$. The base directive $d^b_t$ is routed to the whole-body controller (WBC) through its native kinematic interface. In parallel, the Object-Interaction Layer (OIL) maps the interaction directive $d^i_t$ together with $(r_t, o_t)$ into an upper-body arm joint directive $u_t$.

The WBC combines $d^b_t$ and $u_t$ to produce whole-body joint-level actions that satisfy the commanded base motion while executing the specified upper-body interaction. This design augments an existing WBC with an object-centric command pathway, enabling contact-rich object interaction while preserving the WBC's internal balance and locomotion mechanisms.

\subsection{Whole-Body Controller Interface}
\label{sec:MHC}

The Masked Humanoid Controller (MHC)~\cite{dugar2025learning} is a learned whole-body controller that maps kinematic robot directives to joint position targets for a humanoid robot. The MHC supports multiple directive modes, including locomotion-only, locomotion-manipulation (upper body with locomotion), and whole-body motion tracking. In this work, we use the loco-manipulation mode, which is the natural interface for OIL since it allows the robot to follow base motion commands while simultaneously executing commanded arm motions.

In loco-manipulation mode, we view the MHC as a whole-body policy
\[
\pi^{\mathrm{WBC}}\!\left(r_t,\, [d^b_t, u_t]\right) \rightarrow a_t,
\]
where $r_t$ is robot proprioception, $d^b_t = (\hat{v}_t, \hat{\omega}_t, \hat{b}_t)$ is a base directive for joystick-style velocity and orientation control, and $u_t \in \mathbb{R}^8$ is an arm joint directive for the two arms (four actuators per arm). The output $a_t$ corresponds to joint position targets for all $J=20$ actuated joints, which are tracked by a low-level PD controller.

We treat the MHC as a fixed low-level controller throughout. Its weights are frozen, and we do not modify its architecture or retrain it during OIL learning. The MHC is trained with substantial domain randomization and is robust to disturbances, with demonstrated sim-to-real transfer \cite{dugar2025learning}. As a result, the OIL framework inherits these stability and robustness properties, simplifying the object-interaction learning problem. Finally, OIL is agnostic to the specific WBC used, provided it supports the loco-manipulation interface described above.

\subsection{Object-Interaction Layer}

OIL is a policy that maps the current robot and object state, along with the interaction directive, to an arm joint directive.
\[
u_t = \pi^{\mathrm{OIL}}(r_t, o_t, d^i_t),
\]
where $u_t \in \mathbb{R}^8$ provides joint position targets for the arm actuators and is consumed by the WBC in loco-manipulation mode (see below). We parameterize $\pi^{\mathrm{OIL}}$ as a two-layer LSTM with 64 hidden units per layer followed by a linear decoder.

Composing OIL with the underlying whole-body controller yields the object-interactive controller. Using the loco-manipulation interface described in \textit{Whole-Body Controller Interface} section earlier, the overall policy is
\[
\pi^{\mathrm{OI}}(r_t, o_t, d_t)
     = \pi^{\mathrm{WBC}}\!\Big(r_t,\big[d^b_t,\; \pi^{\mathrm{OIL}}(r_t, o_t, d^i_t)\big]\Big),
\]
where $d_t = (d^i_t, d^b_t)$ is the directive at time $t$. The base directive $d^b_t$ is passed directly to the WBC, while OIL provides the arm directive $u_t$.

\textbf{End-effector-only control via masking.}
During training, we also expose OIL to end-effector-only directives by masking object-related inputs (e.g., object state, object targets, and contact flags). This enables a single OIL policy to support both free-space arm motion and contact-rich interaction through the same interface, without switching controllers. Note that this provides a free-space end-effector command interface on top of the MHC (our chosen WBC), which does not natively support end-effector-space control.

\section{OIL Training}
\label{sec:training}

We train the OIL policy $\pi^{\mathrm{OIL}}$ via reinforcement learning in the MuJoCo simulator with the Digit V3 humanoid model using Proximal Policy Optimization (PPO)~\cite{Schulman2017ProximalPO}. The underlying whole-body controller $\pi^{\mathrm{WBC}}$ (MHC in our experiments) remains frozen throughout, so learning is confined to $\pi^{\mathrm{OIL}}$. Below, we describe the training procedure, including episode sampling, reward design, and domain randomization.

\subsection{Episode Sampling and Curriculum} 
\label{sec:episodes}

Training is episodic. At the start of each episode, we sample a scene instance, including object properties and initial conditions, and then sample a sequence of directive targets from a training distribution. Directives are held fixed for a command window and periodically resampled within the episode to generate diverse behaviors. Episodes terminate early if the robot falls or if object dynamics become unstable. Specifically, we terminate when
\begin{enumerate}
    \item base orientation exceeds $\pm 60^\circ$ in roll or pitch, or
    \item object speed exceeds $3.5\,\mathrm{m/s}$.
\end{enumerate}
Otherwise, episodes run for their scheduled duration. We use a two-stage curriculum that begins with free-space end-effector control and then expands to object interaction.

\textbf{Stage 1: End-effector tracking.}
We train OIL to produce arm joint directives that track commanded end-effector targets in free space. Object-related inputs in $d^i_t$ are masked (e.g., object state, object targets, and contact flags), so OIL conditions only on the hand targets within $d^i_t$. Episodes last $8$~s, with command windows sampled uniformly from $[2,4]$~s. The distribution over directives samples random root directives, within specified limits, and random end-effector commands around the robot that fall within feasible limits and do not overlap with the robot body. Details will be provided in \textit{Appendix A}. 

\textbf{Stage 2: Object interaction.}
In Stage 2, we introduce full object-interaction directives and train OIL to realize desired object motion and contact intent. Episodes last $12$~s and are segmented into command windows. At the start of each window, we sample either an object-interaction instance or an end-effector-only instance (with object-related fields masked) with probabilities $0.8$ and $0.2$, respectively. For object-interaction windows, we sample longer durations $\tau \sim \mathcal{U}(7,10)$~s to allow sustained contact and object motion.

Object-interaction instances are specified as \emph{directive sequences} without task labels, mode indicators, or explicit phase boundaries. Each object-interaction window samples a skill chain
\[
\mathcal{C} = [m_1, m_2, \ldots, m_K],
\]
where each phase $m_i$ is a parameterized directive generator. For example, pick-and-place can be expressed as phases such as alignment, approach, contact, and lift, each producing hand, object, and base targets through the same directive interface. At phase initialization, we sample targets and a duration $\tau_i \sim \mathcal{U}(\tau_{\min}, \tau_{\max})$. The active phase then generates the time-varying directive
\[
d_t = (d^i_t,\; d^b_t),
\]
and execution advances to the next phase when the elapsed time exceeds $\tau_i$. Since the policy observes only the current directive $d_t$ (and its execution history), behaviors such as sliding, lifting, and transport are expressed through the same directive format. Details about phases and tasks are provided in \textit{Appendix B}.

We construct the Stage 2 distribution over interactive tasks from five behaviors chosen to span key axes of non-prehensile whole-body interaction: contact modality (uni- vs.\ bimanual), support condition (table-supported vs.\ lifted), and locomotion coupling (stationary vs.\ moving base). These behaviors require transitions between distinct interaction regimes and encourage generalization to new directive sequences.

\begin{enumerate}
    \item \textbf{Uni-manual Slide:} One randomly selected hand contacts the near face of the object and pushes it forward by $\Delta x \sim \mathcal{U}(0.1,0.3)$\,m along the surface.
    \item \textbf{Bi-manual Slide:} Both hands slide the object to a randomly sampled target location on the same surface.
    \item \textbf{Pick-and-place:} The robot establishes bimanual side contact, lifts the object by $\Delta z \sim \mathcal{U}(0.0,0.6)$\,m, and places it at a new location on the same surface.
    \item \textbf{Loco-transport:} The robot lifts the object and carries it while walking, following commanded velocities $v_x \sim \mathcal{U}(-0.5,0.8)$\,m/s, $v_y \sim \mathcal{U}(-0.3,0.3)$\,m/s, and turn rate $\omega \sim \mathcal{U}(-0.4,0.4)$\,rad/s.
    \item \textbf{Large-box Push:} The robot pushes a large floor-level box that is too wide to lift, maintaining continuous bimanual surface contact while walking forward at $v_x \sim \mathcal{U}(0.5,1.0)$\,m/s.
\end{enumerate}

\textit{Scene Generation.}
For each of the five training behaviors, we define an initial scene distribution over robot pose and object properties (position, size, and weight). Given sampled scene parameters, a hand-coded directive generator produces target trajectories for the robot and the object, along with contact signals. More details are provided in \textit{Appendix C}. All parameters are resampled at the start of each episode, and the scene and directive generators will be made publicly available.

\textbf{Domain Randomization.} To facilitate sim-to-real transfer, we randomize physical parameters at episode initialization. Object properties are sampled uniformly, including mass $m \sim \mathcal{U}(0.1, 3.0)\,\mathrm{kg}$, friction coefficient $\mu \sim \mathcal{U}(0.5, 1.5)$, and box dimensions $\sim \mathcal{U}(0.16, 0.25)\,\mathrm{m}$ per axis. For large boxes, we sample larger dimension ranges. We also randomize the object-state update rate provided to OIL by sampling a frequency $f \sim \mathcal{U}(7, 15)\,\mathrm{Hz}$ and holding $o_t$ constant between updates, simulating perception latency from external tracking systems. Robot dynamics randomization follows the ranges used in MHC training~\cite{dugar2025learning}.

\subsection{Reward Design} 

At each timestep $t$, the total reward $r(t)$ includes five components that capture how well the directive is being followed and important regularization factors for stability and sim-to-real transfer.
\[
r(t) =
r_{\text{hand}}(t)
+ r_{\text{obj}}(t)
+ r_{\text{contact}}(t)
+ r_{\text{reg}}(t)
- r_{\text{coll}}(t)
\]
Each term is only active when the corresponding directive field is unmasked. Below, we describe each component. Further details are in the \textit{Appendix D}. 

\textbf{Hand Tracking ($r_{\text{hand}}$).}
Let $p^h_t \in \mathbb{R}^3$ denote the measured hand position for hand $h \in \{L,R\}$ in the chosen hand frame (consistent with the directive), and let $\hat{p}^h_t$ denote the commanded hand target from $d^i_t$. We use an exponential tracking kernel with an additional proximity bonus:
\begin{small}
\begin{equation*}
\begin{aligned}
r_{\text{hand}}(t)
&=
\sum_{h \in \{L,R\}}
w_{\text{hand}}
\exp\!\left(-s_{\text{hand}} \|p^h_t - \hat{p}^h_t\|_2 \right)
\\
&\quad +
w_{\text{bonus}}
\mathbb{I}\!\left[\|p^h_t - \hat{p}^h_t\|_2 < \epsilon_{\text{hand}}\right].
\end{aligned}
\end{equation*}
\end{small}

\textbf{Object Pose Tracking ($r_{\text{obj}}$).}
Rather than tracking only object position, we track object translation and its planar orientation about the vertical axis. We represent the object pose using two \emph{constellation points}: the centroid $p^0_t \in \mathbb{R}^3$ and a second point $p^1_t \in \mathbb{R}^3$ rigidly attached to the object surface along its local $x$-axis. The directive specifies the corresponding targets $\hat{p}^0_t$ and $\hat{p}^1_t$, which define the desired translation and 2D orientation. More generally, three non-collinear constellation points could be used to track a full $SE(3)$ pose.

We define the object tracking reward as a sum of squared errors over the constellation points, with an additional proximity bonus based on the centroid:
\begin{small}
\begin{equation*}
\begin{aligned}
r_{\text{obj}}(t)
&=
- w_{\text{obj}}
\sum_{k \in \{0,1\}}
\|p^k_t - \hat{p}^k_t\|_2^2
\\
&\quad +
w_{\text{obj,bonus}}
\mathbb{I}\!\left[\|p^0_t - \hat{p}^0_t\|_2 < \epsilon_{\text{obj}}\right].
\end{aligned}
\end{equation*}
\end{small}

\textbf{Contact Matching ($r_{\text{contact}}$).}
Desired hand--object contacts are specified by binary flags $c^h_t \in \{0,1\}$ in $d^i_t$. Let $f^h_t$ denote the magnitude of the measured hand--object contact force, and define
\begin{equation*}
\hat{c}^h_t = \mathbb{I}\!\left[f^h_t > f_{\min}\right].
\end{equation*}
For lift or carry interactions, we additionally require sufficient upward support force for contact to be considered valid:
\begin{equation*}
\hat{c}^h_t = \mathbb{I}\!\left[f^h_t > f_{\min}\right] \cdot \mathbb{I}\!\left[f^{h,\uparrow}_t \ge \alpha\, m g \right],
\end{equation*}
where $m$ is the object mass and $\alpha$ is a constant. The contact reward is then
\begin{equation*}
r_{\text{contact}}(t) =
w_{\text{match}}
\sum_{h \in \{L,R\}}
\mathbb{I}\!\left[\hat{c}^h_t = c^h_t \right].
\end{equation*}

\textbf{Regularization ($r_{\text{reg}}$).}
To discourage high-frequency motion and excessive actuation, we apply penalties on action changes and power consumption:

\begin{small}
\begin{equation*}
\begin{aligned}
r_{\text{reg}}(t) 
&=
w_{\Delta a}\,
\exp\!\left(- s_{\Delta a}\,\|a_t - a_{t-1}\|_1\right)
\\
&\quad +
w_{\text{power}}\,
\exp\!\left(- s_{\text{power}}\,\mathrm{mean}\!\left(|\tau_t \odot \dot{\theta}_t|\right)\right),
\end{aligned}
\end{equation*}
\end{small}
where $a_t$ are the WBC joint targets produced by $\pi^{\mathrm{OI}}$, $\tau_t$ are actuator torques, and $\dot{\theta}_t$ are joint velocities.

\textbf{Collision Penalty ($r_{\text{coll}}$).}
We penalize unsafe torso-environment collisions: 
\[
r_{\mathrm{coll}}(t) = w_{\mathrm{coll}}\; \mathbb{I}\!\left[\text{torso collision at } t\right].
\]

\section{Results}
\label{sec:results}
We evaluate OIL along three axes that correspond to its design claims: (i) whether a single policy with object-level commands improves over robot-only control and matches task-specific policies trained per behavior, (ii) whether the OIL controller generalizes to object properties outside its training distribution, and (iii) whether a single controller supports zero-shot composition of novel behavior sequences. 

\begin{table}[t]
  \centering
  \caption{%
    Success rate (\%, $\uparrow$) and object position error
    (cm, $\downarrow$) on five tasks. Best result per task is \textbf{bolded}. NoObj denotes $\text{MHC}_{\text{NoObj}}$; OIL and NoObj use a single controller, whereas TaskSp uses a different controller specifically trained for each task.%
  }
  \label{tab:results}
  \setlength{\tabcolsep}{3pt}
  \renewcommand{\arraystretch}{1.05}
  \begin{tabular}{ll ccc}
    \toprule
    Task & & NoObj & OIL \textit{(ours)} & TaskSp. \\
    \midrule
    \multirow{2}{*}{Uni Slide}
      & $\text{Succ.}$\,(\%)           & 58.1 & 89.1           & \textbf{90.0} \\
      & $\text{Obj}_{\text{err}}$\,(cm) & 3.83 & \textbf{2.35}  & 3.53          \\
    \midrule
    \multirow{2}{*}{Bi Slide}
      & $\text{Succ.}$\,(\%)           & 99.4 & \textbf{100.0} & 99.0          \\
      & $\text{Obj}_{\text{err}}$\,(cm) & 2.46 & \textbf{1.30}  & 2.30          \\
    \midrule
    \multirow{2}{*}{Pick \& Place}
      & $\text{Succ.}$\,(\%)           & 63.6 & 94.3           & \textbf{94.5} \\
      & $\text{Obj}_{\text{err}}$\,(cm) & 3.21 & \textbf{2.23}  & 3.09          \\
    \midrule
    \multirow{2}{*}{Loco-manip.}
      & $\text{Succ.}$\,(\%)           & 15.8 & \textbf{84.1}  & 81.4          \\
      & $\text{Obj}_{\text{err}}$\,(cm) & 4.08 & \textbf{2.53}  & 3.15          \\
    \midrule
    \multirow{2}{*}{Large-Box Push}
      & $\text{Succ.}$\,(\%)           & 18.3 & \textbf{84.6}  & 76.9          \\
      & $\text{Obj}_{\text{err}}$\,(cm) & 5.96 & 3.01           & \textbf{1.84} \\
    \midrule
    \multirow{2}{*}{\textbf{Mean}}
      & $\text{Succ.}$\,(\%)           & 51.0 & \textbf{90.4}  & 88.4          \\
      & $\text{Obj}_{\text{err}}$\,(cm) & 3.93 & \textbf{2.28}  & 2.76          \\
    \bottomrule
  \end{tabular}
\end{table}


\subsection{Evaluation Setup and Metrics:}
We evaluate OIL on Digit V3 in MuJoCo using \emph{held-out} rollouts with newly sampled scenes and directives from the five task distributions described in \textit{Episode Sampling and Curriculum} section above. This in-distribution evaluation is necessary to compare against \emph{task-specific} controllers, which must be trained and evaluated on the same task family. All results are computed on fresh random seeds and scene instances that were not replayed from training. We then evaluate OIL under out-of-distribution (OOD) scene properties and novel behavior compositions to study generalization beyond the training distributions. All simulation evaluations are averaged over 5000 trials per task.

\textbf{Metrics.} 
We report success rate ($\text{Succ.}$, \%, $\uparrow$) and mean object tracking error ($\text{Obj}_{\text{err}}$, cm, $\downarrow$). An episode is successful if the final object position is within a threshold of the last commanded target: 5\,cm for stationary tasks and 15\,cm for locomotion-coupled tasks, reflecting the larger positional variance inherent to moving-base interactions. Object tracking error is the mean per-step 
Euclidean distance between the actual and commanded object position in the robot torso frame, averaged over all interaction steps in an episode. Together, these metrics capture both task completion and the quality of object motion throughout execution.
\begin{figure*}[t]
    \centering
    \begin{minipage}[t]{0.48\textwidth}
        \centering
        \includegraphics[width=\linewidth]{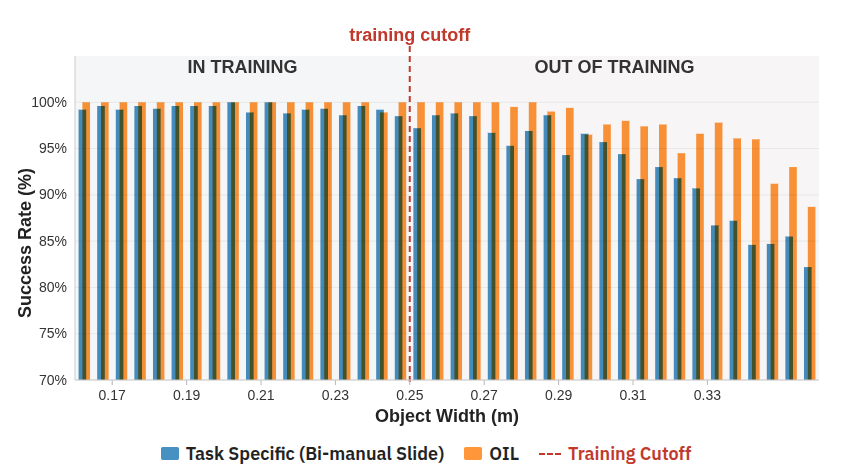}
    \end{minipage}
    \hfill
    \begin{minipage}[t]{0.48\textwidth}
        \centering
        \includegraphics[width=\linewidth]{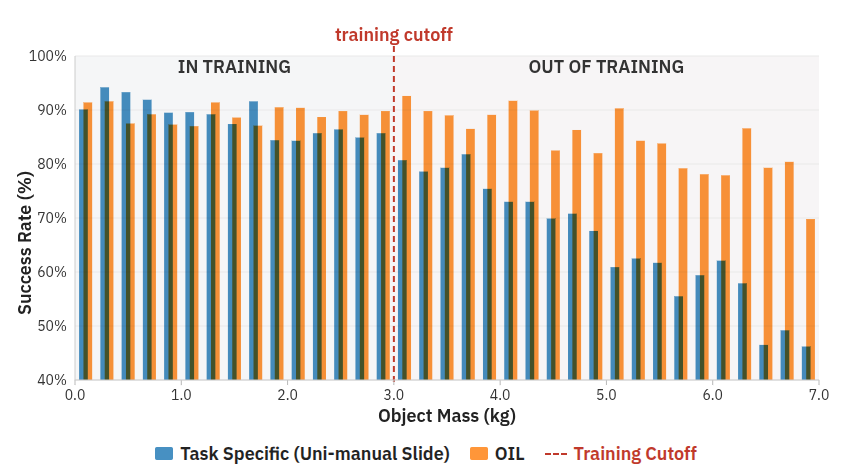}
    \end{minipage}

    \caption{Out-of-distribution robustness to object properties for OIL and task-specific policies. Left: success rate on bi-manual slide as object width increases beyond the training range. Right: success rate on uni-manual slide as object mass increases beyond the training range. Dashed lines indicate training cutoffs.}
    \label{fig:ood_properties}
\end{figure*}
\subsection{Baseline Comparison}
\label{sec:results_baselines}

We compare OIL against two baselines: (1) \textbf{$\text{MHC}_{\text{NoObj}}$}, the underlying whole-body controller driven by end-effector directives without object state, and (2) \textbf{Task-specific} policies trained independently per task with full access to object state, using the same frozen MHC base and network architecture as OIL. OIL is evaluated as a single controller across 
all five tasks without per-task retraining.

Table~\ref{tab:results} reports results across all five tasks. OIL substantially outperforms $\text{MHC}_{\text{NoObj}}$, with mean success 
rate improving from 51.0\% to 90.4\% and mean object tracking error reducing from 3.93\,cm to 2.28\,cm. The $\text{MHC}_{\text{NoObj}}$ attains a success rate comparable to OIL on only the bi-manual slide task, which is the task where the object motion is most constrained by the robot motion itself. The gains observed for OIL are largest on non-stationary tasks where object intent cannot be reliably inferred from robot kinematics alone: loco-transport ($15.8\% \to 84.1\%$) and large-box push ($18.3\% \to 84.6\%$). This confirms the benefit of providing a controller with directives that include explicit object interaction information. 

OIL further matches task-specific policies on mean success rate (90.4\% vs.\ 88.4\%) while achieving lower mean object tracking error (2.28\,cm vs.\ 
2.76\,cm). These results show that object-level commands are necessary for robust interaction and that a single policy with such commands can match per-task engineering without task-specific fine-tuning.

\subsection{Generalization to Out-of-Distribution Object Properties}

Figure~\ref{fig:ood_properties} evaluates robustness to object properties outside the training distribution. We focus on two representative settings: \textit{Bi-manual Slide} for width sweeps and \textit{Uni-manual Slide} for mass sweeps. These tasks were chosen because, within the training range, OIL and the corresponding task-specific policy achieve nearly identical success rates, making the out-of-distribution comparison more diagnostic. We sweep object width and mass beyond the training cutoffs of 0.25\,m and 3.0\,kg, respectively, and compare OIL against a task-specific policy trained for each task.

Within the training range, both methods perform comparably. Beyond the training range, OIL generalizes substantially better. In the width sweep (Figure~\ref{fig:ood_properties}, left), OIL maintains above $\sim$95\% success well past 0.25\,m, while the task-specific policy drops to $\sim$85\%. In the mass sweep (Figure~\ref{fig:ood_properties}, right), the task-specific policy falls below $\sim$50\% at 6.5\,kg, whereas OIL remains above $\sim$80\%.

We attribute this improvement to OIL's multi-behavior training distribution and directive-driven conditioning. Training across diverse robot and object motions encourages strategies that remain effective when object dynamics shift and the object deviates from its nominal trajectory. In contrast, task-specific policies are optimized for a narrower set of motion patterns and degrade more quickly under out-of-distribution properties that induce trajectory deviations.

\begin{figure}[t]
  \centering
  \includegraphics[width=\columnwidth,height=0.7\columnwidth]{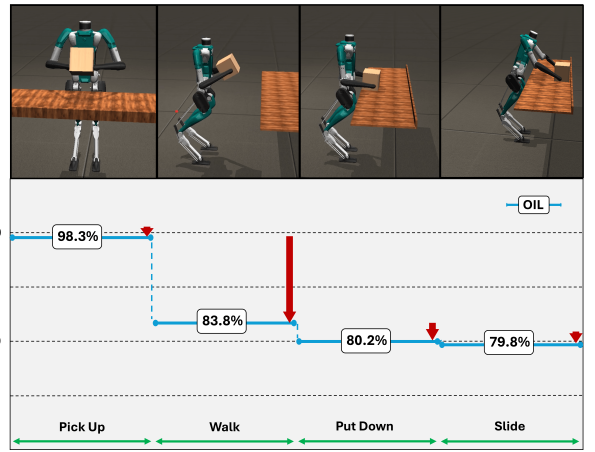}
  \caption{Success rate across task composition (Pick-Up → Walk → Put Down → Slide). Red arrows indicate performance drops at task switches (no policy switch); green arrows denote task segments.}
  \label{fig:compose}
\end{figure}

\subsection{Generalization to Novel Behavior Compositions}

A key property of OIL's directive-based design is that novel behavior sequences can be constructed at test time by chaining directive phases, without retraining. We evaluate this by constructing a four-stage composed task: pickup $\rightarrow$ loco-transport $\rightarrow$ place $\rightarrow$ bi-manual slide. We evaluate a single OIL policy on 10{,}000 episodes without fine-tuning, with object mass, size, initial position, walking velocity, and placement locations randomized using the training ranges.

Figure~\ref{fig:compose} reports per-stage success rates. Performance drops are modest at most transitions (1.7\% at pick-up, 3.6\% at place, 1.3\% at slide), with the largest drop at loco-transport (14.5\%), where errors from the preceding pick phase compound with the demands of locomotion-coupled holding. The final success rate of 78.9\% across the full four-stage sequence demonstrates that OIL's command interface supports zero-shot composition and gracefully degrades across stages. The key failure mode for transitioning to loco-transport was the box slipping out of the pinch grasp for box mass $\leq$ 2.5 kg. 

\subsection{Sim-to-Real Transfer}
We deploy OIL zero-shot on the Digit V3 humanoid, with the policy running 
at 50\,Hz and object state provided at 10\,Hz. Object and torso poses are 
estimated in a shared world frame using an overhead RealSense D455 camera with 
ArUco markers, then transformed into the robot torso frame before being 
passed to OIL. To mitigate tracking degradation due to marker jitter and intermittent occlusions caused by the robot's safety harness, we employ a Kalman filter that retains 50 time steps of history to enable robust re-initialization upon occlusion recovery \cite{tony2020vision}. Between perception updates, the last known object state is held constant, matching the randomized update rate used during training. Directives are generated either manually by an operator or via a finite-state machine that operates purely at the command interface level, producing no joint trajectories or contact forces.

OIL transfers successfully to hardware without any modification, executing 
uni- and bi-manual sliding, pick-and-place, and loco-transport. These results confirm that the sim-to-real gap is well bridged by the domain randomization applied during training, and that the directive interface is sufficiently general to support real-world deployment across diverse interaction behaviors. Video demonstrations of these behaviors are provided in \textit{Appendix~E}.

\subsection{Limitations}
Despite strong performance, OIL exhibits two primary failure modes. In the uni-manual slide, success is sensitive to the lateral accuracy of the hand contact point. When the end effector contacts the object off-center, the resulting torque causes the object to rotate rather than translate. While OIL can partially compensate through object-state feedback, large deviations exceed what continuous arm adjustments can correct, and open-loop directive following prevents a corrective reapproach. A second failure mode occurs in bimanual tasks, where the policy relies on the symmetric side contacts to support the object. Insufficient upward tangential force can cause the object to slip out of the grasp, particularly for heavier or smoother objects. In both cases, the underlying issue is that once contact quality degrades beyond a threshold, OIL has no mechanism to recover it. This points to the important direction of incorporating contact and/or low-resolution force sensors into the hardware and OIL observation space.

\section{FUTURE WORK}
Future work will expand the interaction directives to include full SE(3) object orientation, different grasp modes, and force/contact targets when this information is available upstream. Another direction is to generate directives in a closed loop, allowing the system to replan when contact fails or degrades instead of being limited to a single attempt. We also plan to incorporate visual observations as input for object sensing, reducing reliance on externally provided object state estimates. 

\section{CONCLUSIONS}
We introduced the Object Interaction Layer (OIL), a simple extension that augments an existing whole-body controller with an object-interactive command interface. We focused on non-prehensile, non-grasping end-effector loco-manipulation and presented a training distribution spanning diverse interaction regimes. Experiments in simulation show that OIL outperforms whole-body control driven only by robot-motion directives, while matching or exceeding the performance of task-specific controllers trained per task. We further show that OIL generalizes to object properties outside the training distribution and to novel directive sequences that compose learned behaviors. Finally, we demonstrate sim-to-real transfer on the Digit V3 robot. Promising directions for future work include extending the interface to support grasping and dexterous end-effectors, contact and force sensors, and incorporating collision-avoidance objectives for non-interactive environment elements.

\section{Acknowledgments}
This work is supported by NSF Award 2321851, DARPA contract HR0011-24-9-0423, and the NVIDIA Academic Grant Program.

\bibliography{references}


\newpage
\section{Appendix A: Directive Sampling}
Hand and locomotion targets follow the distributions in Table~\ref{tab:directive-sampling}. Hand targets near the torso are rejected. Stationary phases command zero velocity; transport and large-box pushing use sampled velocities. Object inputs remain masked during free-space training.

\begin{table}[ht]
\centering
\caption{Directive sampling distributions. Hand coordinates are
relative to the base; velocity ranges apply during locomotion phases.}
\small
\setlength{\tabcolsep}{3pt}
\begin{tabular}{@{}lr@{}}
\toprule
Directive & Distribution \\
\midrule
Hand $x$, both hands & $\mathcal U(0,0.6)$ m \\
Hand $y$, left & $\mathcal U(-0.3,0.7)$ m \\
Hand $y$, right & $\mathcal U(-0.7,0.3)$ m \\
Hand $z$, both hands & $\mathcal U(-0.1,0.9)$ m \\
Base height & $\mathcal U(0.36,1.05)$ m \\
Base height, large-box & $\mathcal U(0.4,0.75)$ m \\
Transport $v_x$ & $\mathcal U(-0.5,0.8)$ m/s \\
Transport $v_y$ & $\mathcal U(-0.3,0.3)$ m/s \\
Transport $\omega$ & $\mathcal U(-0.4,0.4)$ rad/s \\
Large-box push $v_x$ & $\mathcal U(0.5,1.0)$ m/s \\
Stationary $(v_x,v_y,\omega)$ & $(0,0,0)$ \\
\bottomrule
\end{tabular}
\label{tab:directive-sampling}
\end{table}

\section{Appendix B: Directive Sequences}
Behaviors combine the phases in Table~\ref{tab:directive-phases}.
Each phase samples targets and a duration, interpolating motion
commands where needed. Transitions are time-based; the policy
receives directives without phase labels.
Table~\ref{tab:behavior-sequences} summarizes the main phase sequence
for each behavior.

\begin{table}[ht]
\centering
\caption{Directive phases used to compose interaction behaviors.}
\small
\setlength{\tabcolsep}{3pt}
\begin{tabular}{@{}p{0.20\columnwidth}p{\dimexpr0.80\columnwidth-6pt\relax}@{}}
\toprule
Phase & Description \\
\midrule
Align & Set base posture and prepare the hands. \\
Approach & Move hands to pre-contact positions. \\
Contact & Establish contact while holding the object target. \\
Slide & Translate the object along its supporting surface. \\
Lift & Raise the object with bilateral contact. \\
Hold & Maintain the commanded object position. \\
Walk & Carry the object while following velocity commands. \\
Stop & Return to standing while supporting the object. \\
Place & Lower the object onto a supporting surface. \\
Walk-push & Push forward with bilateral contact while walking. \\
\bottomrule
\end{tabular}
\label{tab:directive-phases}
\end{table}

\begin{table}[ht]
\centering
\caption{Main directive phases for each interaction behavior.}
\small
\setlength{\tabcolsep}{3pt}
\begin{tabular}{@{}p{0.30\columnwidth}p{\dimexpr0.70\columnwidth-6pt\relax}@{}}
\toprule
Behavior & Main phases \\
\midrule
Uni-manual slide & Align $\rightarrow$ approach $\rightarrow$ contact $\rightarrow$ slide \\
Bi-manual slide & Align $\rightarrow$ approach $\rightarrow$ bilateral contact $\rightarrow$ slide \\
Pick-and-place & Align $\rightarrow$ contact $\rightarrow$ lift $\rightarrow$ hold $\rightarrow$ place \\
Loco-transport & Align $\rightarrow$ contact $\rightarrow$ lift $\rightarrow$ walk $\rightarrow$ stop $\rightarrow$ place \\
Large-box push & Align $\rightarrow$ approach $\rightarrow$ bilateral contact $\rightarrow$ walk-push \\
\bottomrule
\end{tabular}
\label{tab:behavior-sequences}
\end{table}

\section{Appendix C: Scene Generation}
Object geometry and placement are sampled within the ranges in
Table~\ref{tab:scene-sampling}. Table height is coupled to commanded
base height $h_b$. Objects start on their supporting surface, with
lateral placement conditioned on the selected hand. Mass, friction,
and observation updates follow the main paper's domain randomization.

\begin{table}[ht]
\centering
\caption{Scene sampling ranges. Large boxes rest on the floor.
Initial table heights span 0.01--1.15 m through the coupled base-height
and offset distributions.}
\small
\setlength{\tabcolsep}{3pt}
\begin{tabular}{@{}lrr@{}}
\toprule
Parameter (m) & Standard & Large box \\
\midrule
Full size $x$ & $[0.16,0.25]$ & $[1.1,1.3]$ \\
Full size $y$ & $[0.16,0.25]$ & $[0.6,1.0]$ \\
Full size $z$ & $[0.16,0.25]$ & $[0.9,1.3]$ \\
Forward distance & $[0.3,0.4]$ & $[0.8,0.9]$ \\
Lateral offset & $[-0.1,0.1]$ & $[-0.1,0.1]$ \\
Initial $h_{\mathrm{table}}-h_b$ & $[-0.35,0.10]$ & --- \\
Placement $h_{\mathrm{table}}-h_b$ & $[-0.10,0.10]$ & --- \\
\bottomrule
\end{tabular}
\label{tab:scene-sampling}
\end{table}

\section{Appendix D: Reward Weights}
The main paper's \textit{Reward Design} subsection describes the hand-tracking, object-tracking, contact-matching, regularization, and collision components of the reward. Table~\ref{tab:oil-training-rewards} summarizes their training weights, with tracking and proximity bonuses listed separately. Regularization is composed of action smoothness and arm power consumption.
\begin{table}[h]
\centering
\caption{Training reward weights, without normalization; undesired
contact is subtracted from the total reward.}
\begin{tabular}{lr}
\toprule
Reward term & Weight \\
\midrule
Hand tracking (per coordinate) & 0.1 \\
Hand proximity bonus & 1.0 \\
Object position tracking & 0.2 \\
Object proximity bonus & 0.2 \\
Contact matching (per hand) & 0.2 \\
Collision Penalty & 0.5 \\
Reg - Action smoothness & 0.1 \\
Reg - Arm power consumption & 0.1 \\
\bottomrule
\end{tabular}
\label{tab:oil-training-rewards}
\end{table}

\section{Appendix E: Sim-to-Real Demonstrations}
The accompanying video demonstrates zero-shot sim-to-real transfer
of OIL across multiple object-interaction behaviors on the Digit V3
humanoid robot:
\url{https://drive.google.com/file/d/1RI56LUpW_dM1pZJrmNrjPz3uODhO5HKy/view}

\end{document}